\pdfoutput=1

\documentclass[11pt]{article}

\usepackage{acl}

\usepackage{times}
\usepackage{latexsym}
\usepackage{xcolor}
\usepackage{enumitem}

\usepackage[T1]{fontenc}

\usepackage[utf8]{inputenc}

\usepackage{microtype}

\title{An Empirical Study on Relation Extraction in the Biomedical Domain}

\author{Yongkang Li \\
Peking University\\
  \texttt{liyongkang@pku.edu.cn}}

\begin{document}
\maketitle
\begin{abstract}
Relation extraction is a fundamental problem in natural language processing. Most existing models are defined for relation extraction in the general domain. However, their performance on specific domains (e.g., biomedicine) is yet unclear. To fill this gap, this paper carries out an empirical study on relation extraction in biomedical research articles. Specifically, we consider both sentence-level and document-level relation extraction, and run a few state-of-the-art methods on several benchmark datasets. Our results show that (1) current document-level relation extraction methods have strong generalization ability; (2) existing methods require a large amount of labeled data for model fine-tuning in biomedicine. Our observations may inspire people in this field to develop more effective models for biomedical relation extraction.
\end{abstract}

\section{Introduction}

Relation extraction, which aims to extract relations of entities, is a fundamental problem in natural language processing. Formally, there are two different settings for relation extraction. Sentence-level relation extraction~\cite{yao2019docred} focuses on each sentence, and the goal is to predict the relation of two entity mentions in a sentence. In contrast to that, document-level relation extraction~\cite{yao2019docred} considers longer texts (e.g., paragraphs, documents) to predict entity relations. Relation extraction can benefit a variety of applications, including knowledge graph construction~\cite{cowie1996information} and question answering~\cite{lin2019kagnet}.

In literature, a variety of methods have been proposed for relation extraction. Classical methods build models with Convolutional Neural Networks (CNN)~\cite{DBLP:journals/corr/Kim14f}, Recurrent Neural Networks (RNN)~\cite{gupta2016table}, and Long Short-Term Memory (LSTM) networks~\cite{zhang2017position}. Despite the good performance, their performance highly depends on the quality of labeled sentences. When labeled sentences are insufficient as in most cases, these methods often suffer from poor results. Recently, with the great success of pre-trained language models (e.g., BERT)~\cite{devlin2018bert,liu2019roberta}, there is a growing interest in approaching relation extraction by fine-tuning pre-trained language models with a few labeled sentences~\cite{han2018fewrel,gao2019fewrel}. Owing to the high capacity of pre-trained language models, these models achieve impressive results on both sentence-level and document-level relation extraction.

However, most existing studies develop relation extraction models for general domains. In general domains, we can easily collect a large amount of data for pre-training language models. Also, annotating relations of entities is relatively easy, which allows us to have many labeled sentences. Nevertheless, for some specific domains (e.g., biomedicine), the available text data is much more limited, and thus it is generally more difficult to train a powerful language model. Moreover, annotating labeled data in specific domains often requires domain experts, which is highly expensive, and thus labeled data is often insufficient in specific domains. Therefore, whether existing relation extraction models are still able to perform well on these specific domains remains unclear and under-explored.

In this paper, we fill this gap by applying these relation extraction models to the biomedical domain, as biomedical relation extraction is attracting growing interest, especially during the COVID-19 pandemic, which enables researchers to quickly identify important knowledge in research papers. To give readers a comprehensive sense of biomedical relation extraction, we focus on both sentence-level and document-level relation extraction. For each task, we choose two benchmark datasets, which are DDI~\cite{ddi}, Chemprot~\cite{chemprot} for sentence-level RE and CDR~\cite{CDR},GDA~\cite{GDA} for ducument-level RE. Also, a few state-of-the-art models are selected for comparison. For sentence-level relation extraction, we choose MTB (Matching the blanks)~\cite{mtb}, TEMP (Typed Entity Marker)~\cite{TEMP}, and the Multitask learning framework~\cite{NLLIE}. For document-level relation extraction, we choose MTB, ATLOP (Adaptive
Thresholding Localized cOntext Pooling)~\cite{ATLOP}, and DocUNet (Document U-shaped Network)~\cite{DOCUNET}. Our results show:
\begin{itemize}[leftmargin=*]
    \item Current methods which have been proved efficient on TACRED/DOCRED will need a large amount of training data to fine-tune.
    \item The framework of current sentence-level models only have limited contributions compared with transformer-based encoders. 
    \item Existing methods for document-level relation extraction are quite strong in terms of generalization ability.
\end{itemize}
We believe our results could inspire future studies on biomedical relation extraction.

\section{Related Work}

Relation extraction is an important problem, which aims at predicting the relation of two entities. Early works of relation extraction focus on sentence-level extraction~\cite{mintz2009distant}, and the goal is to predict the relation of two entities in a sentence. However, in many cases, the relation of two entities is described across multiple sentences, and sentence-level relation extraction methods cannot deal with these cases. Inspired by the observation, there are also some recent studies focusing on document-level relation extraction~\cite{yao2019docred}, aiming to predict entity relations in longer text, such as paragraphs and documents.

In terms of methodology, classical relation extraction methods apply classification algorithms (e.g., SVM) to bag-of-words features for relation extraction~\cite{zelenko2003kernel}, but they often suffer from poor performance as they cannot deal with word orders. Later on, many studies apply deep neural networks for relation extraction, including convolutional neural networks~\cite{DBLP:journals/corr/Kim14f}, recurrent neural networks~\cite{gupta2016table}, and long-short term memory neural networks~\cite{zhang2017position}. As these methods are able to take word orders into consideration, they get significant improvements over traditional relation extraction methods. However, these methods often require a large amount of labeled data to train effective models. On many datasets where the labeled data is limited, their performance is far from satisfactory. The recent success of pre-trained language models~\cite{devlin2018bert,liu2019roberta} opens a door for solving the problem. By pre-training neural language models (e.g., Transformers~\cite{vaswani2017attention}) on a huge amount of unlabeled data, these models can effectively capture the semantics of texts. With such capabilities, only a few labeled sentences are sufficient to fine-tune these models for relation extraction~\cite{han2018fewrel,gao2019fewrel}. Hence, these methods achieve impressive results on many relation extraction datasets. Despite the success, whether these methods can also achieve good results on specific domains is still unclear. In this paper, we address the issue by conducting systematical experiments on relation extraction in the biomedical domain. Based on the results, we further highlight a few insights and guidance for biomedical relation extraction, which would benefit futures studies in this field.

\section{Settings}

In this paper, we consider two settings of relation extraction.

\begin{itemize}[leftmargin=*]
    \item Sentence-level Relation Extraction. The first setting focuses on sentence-level extraction. Specifically, given two entities in a sentence, we aim to predict their relation, which is either a relation in the given vocabulary, or a special relation called ``no relation''. This setting mainly targets at simple relations between entities, which can be described within single sentences.
    \item Document-level Relation Extraction. The second setting studies document-level extraction. Specifically, we consider entities which are mentioned in longer texts, such as paragraphs or documents. The goal is to still to predict the relation of each pair of entities. The document-level setting mainly focuses on more complicated relations, which cannot be mentioned in each individual sentence.
\end{itemize}

\section{Model}

In this study, we choose a few state-of-the-art methods on benchmark datasets (e.g., TacRED~\cite{zhang2017position} and DocRED~\cite{yao2019docred}) for comparison.

For sentence-level relation-extraction, there are few work tested their performance on biomedical data. We evaluate their performance on their original settings with BioBERT~\cite{biobert} as their encoder. The models are listed below:

\begin{itemize}[leftmargin=*]
\item\textbf{MTB (Matching the Blanks)}~\cite{mtb} takes pairs of blank-containing relation statements as input for prediction, and uses an objective that encourages relation representations to be similar if they range over the same pairs of entities.

\item \textbf{TEMP (Typed Entity Marker (punct))}~\cite{TEMP} is a variant of the typed entity marker technique that marks the entity span and entity types without introducing new special tokens. This leads to promising improvement over existing RE models on TACRED.

\item \textbf{NLLIE}~\cite{NLLIE} consists of several neural models with identical structures but different parameter initialization. These models are jointly optimized with the task-specific losses and are regularized to generate similar predictions based on an agreement loss, which prevents overfitting on noisy labels.
\end{itemize}

For document-level relation-extraction, we have selected two transformer-based model. Existing work has explored the performance of these methods with SciBERT~\cite{scibert}. To better evaluate their performance on biomedical data, we further evaluate ATLOP, DocUNet with BioBERT as their Encoder. The model are listed below:

\begin{itemize}[leftmargin=*]

\item \textbf{ATLOP}~\cite{ATLOP} mainly consists of two techniques, adaptive thresholding and localized context pooling. The adaptive thresholding replaces the global threshold for multi-label classification in the prior work with a learnable entities-dependent threshold. The localized context pooling directly transfers attention from pre-trained language models to locate relevant context that is useful to decide the relation.

\item \textbf{DocUNet}~\cite{DOCUNET} approaches the problem by predicting an entity-level relation matrix to capture local and global information, parallel to the semantic segmentation task in computer vision. Specifically, it leverage an encoder module to capture the context information of entities and a segmentation module over the image-style feature map to capture global interdependency among triples.
\end{itemize}

\section{Experiment}

\subsection{Dataset} 

We evaluated multiple models on document-level and sentence-level RE datasets. The data statistics are listed in Table~\ref{tab:stats1} (sentence-level) and Table~\ref{tab:stats2} (document-level).

\begin{table}[!htb]
\caption{Statistics of sentence-level RE datasets.}
\label{tab:stats1}
\centering
\begin{tabular}{c c c}
\hline  
 \textbf{Statistics}\textbackslash{}\textbf{Dataset}&\textbf{DDI}&\textbf{ChemProt}\\
\hline  
Train&25296&18035\\
Dev&2496&11268\\
Test&5716&15645\\
Relations&5&6\\
\hline 
\end{tabular}
\end{table}

\begin{itemize}[leftmargin=*]
    \item \textbf{ChemProt}~\cite{chemprot} consists of 1,820 PubMed abstracts with chemical-protein relations. We evaluate on six classes: CPR:3, CPR:4, CPR:5, CPR:6, and CPR:9 and no-relation.
    \item \textbf{DDI}~\cite{ddi} is a collection of 792 texts selected from the  DrugBank database and other 233 Medline abstracts. We evaluate five types: DDI-effect, DDI-int, DDI-mechanism, DDI-advise as well as no-relation.
\end{itemize}

\begin{table}[!htb]
\caption{Statistics of document-level RE datasets.}
\label{tab:stats2}
\centering
\begin{tabular}{c c c}
\hline 
\textbf{Statistics}\textbackslash{}\textbf{Dataset} & \textbf{CDR} & \textbf{GDA}   \\\hline 
Train                             & 500 & 23353 \\
Dev                               & 500 & 5839  \\
Test                              & 500 & 1000  \\
Relations                         & 2   & 2     \\
Avg. Entities per Doc.            & 7.6 & 5.4   \\
Avg. Mention per Entity           & 2.7 & 3.3   \\\hline 
\end{tabular}
\end{table}

\begin{itemize}[leftmargin=*]
    \item\textbf{CDR}~\cite{CDR} is a relation extraction dataset in the biomedical domain with 500 training samples.  It is aimed to extract the relations between chemical and disease.
    \item\textbf{GDA}~\cite{GDA} consists of 23,353 training samples. It is aimed to predict the relations between genes and disease.
\end{itemize}

\subsection{Experimental Settings}  

Our model was implemented based on Pytorch. We used cased Bio-BERT-base, and RoBERTa-large as the encoder on DDI and  ChemProt and cased BioBERT-base and SciBERT-base on CDR and GDA. We tuned the hyperparameters on the development set. and evaluated our model with micro F1.

\subsection{Results}

The results of sentence-level and document level relation extraction are presented in Table~\ref{tab:results1} and Table~\ref{tab:results2} respectively. For sentence-level RE, the results of SciBERT and BioBERT are from a former work~\cite{DOCUNET}. For document-level RE, the results of ATLOP-SciBERT and DocUNet-SciBERT are taken from~\citet{DOCUNET}.  


\begin{table}[!htb]
\caption{Results of sentence-level RE.}
\label{tab:results1}
\centering
\begin{tabular}{c c c}
    \hline  
    \textbf{Models} & \textbf{DDI}& \textbf{Chemprot}\\
    \hline
    BioBERT&80.88&76.14\\
    SciBERT&81.22&75\\
    BioBERT-MTB&90.45&76.34\\
    BioBERT-NLLIE&92.27&76.97\\
    BioBERT-TEMP&93.30&76.75\\
    \hline 
\end{tabular}
\end{table}
From the results, we see that the imrpovement of state-of-the-art models on Chemprot is very limited, whereas the improvement on DDI is relatively significant. The underlying reason might be that Chemprot is extracted from PubMed where BioBERT has already been trained, which implies that current methods only have limited contribution compared to the encoder. However, on DDI, as the encoder still has a gap to the unseen corpus, current methods work pretty well. In conclusion, the generalization ability of those sentence-level methods is not satisfactory.

\begin{table}[!htb]
\caption{Results of document-level RE.}
\label{tab:results2}
\centering
\begin{tabular}{ c c c }
\hline  
\textbf{Models}& \textbf{CDR}& \textbf{GDA}\\
\hline 
ATLOP-SciBERT&69.4&82.5\\
DocUNet-SciBERT&76.3&85.3\\
BioBERT-MTB&74.23&79.61\\
ATLOP-BioBERT&77.10&90.20\\
DocUNet-BioBERT&78.37&90.37\\
\hline 
\end{tabular}
\end{table}

Document-level RE is a more challenging task. However, the improvement on CDR and GDA is very obvious and state-of-the-art models turn out to have a relatively powerful ability of denoising. Compared with CDR, methods achieves a higher improvement on GDA. This is as expected, because GDA has a great amount of training data than CDR. Both DocUNet and ATLOP are originally proposed and designed for the DocRED Relation Extraction task which has 101,873 documents.

\section{Conclusion}

This paper conducts an empirical analysis of relation extraction in biomedicine. We consider both sentence-level and document-level relation extraction. For each task, a few state-of-the-art models are considered. Our results show that (1) existing methods for document-level relation extraction are quite strong in terms of generalization ability; (2) existing methods still require a large amount of labeled data for training effective models. Therefore, we believe that few-shot learning techniques, which are able to reduce the reliance of models on labeled data, are important to explore to improve biomedical relation extraction.

\bibliography{anthology}

\begin{thebibliography}{24}
\expandafter\ifx\csname natexlab\endcsname\relax\def\natexlab#1{#1}\fi

\bibitem[{Beltagy et~al.(2019)Beltagy, Lo, and Cohan}]{scibert}
Iz~Beltagy, Kyle Lo, and Arman Cohan. 2019.
\newblock Scibert: A pretrained language model for scientific text.
\newblock \emph{arXiv preprint arXiv:1903.10676}.

\bibitem[{Cowie and Lehnert(1996)}]{cowie1996information}
Jim Cowie and Wendy Lehnert. 1996.
\newblock Information extraction.
\newblock \emph{Communications of the ACM}, 39(1):80--91.

\bibitem[{Devlin et~al.(2018)Devlin, Chang, Lee, and
  Toutanova}]{devlin2018bert}
Jacob Devlin, Ming-Wei Chang, Kenton Lee, and Kristina Toutanova. 2018.
\newblock Bert: Pre-training of deep bidirectional transformers for language
  understanding.
\newblock \emph{arXiv preprint arXiv:1810.04805}.

\bibitem[{Gao et~al.(2019)Gao, Han, Zhu, Liu, Li, Sun, and
  Zhou}]{gao2019fewrel}
Tianyu Gao, Xu~Han, Hao Zhu, Zhiyuan Liu, Peng Li, Maosong Sun, and Jie Zhou.
  2019.
\newblock Fewrel 2.0: Towards more challenging few-shot relation
  classification.
\newblock \emph{arXiv preprint arXiv:1910.07124}.

\bibitem[{Gupta et~al.(2016)Gupta, Sch{\"u}tze, and Andrassy}]{gupta2016table}
Pankaj Gupta, Hinrich Sch{\"u}tze, and Bernt Andrassy. 2016.
\newblock Table filling multi-task recurrent neural network for joint entity
  and relation extraction.
\newblock In \emph{Proceedings of COLING 2016, the 26th International
  Conference on Computational Linguistics: Technical Papers}, pages 2537--2547.

\bibitem[{Han et~al.(2018)Han, Zhu, Yu, Wang, Yao, Liu, and
  Sun}]{han2018fewrel}
Xu~Han, Hao Zhu, Pengfei Yu, Ziyun Wang, Yuan Yao, Zhiyuan Liu, and Maosong
  Sun. 2018.
\newblock Fewrel: A large-scale supervised few-shot relation classification
  dataset with state-of-the-art evaluation.
\newblock \emph{arXiv preprint arXiv:1810.10147}.

\bibitem[{Herrero-Zazo et~al.(2013)Herrero-Zazo, Segura-Bedmar, Mart{\'\i}nez,
  and Declerck}]{ddi}
Mar{\'\i}a Herrero-Zazo, Isabel Segura-Bedmar, Paloma Mart{\'\i}nez, and
  Thierry Declerck. 2013.
\newblock The ddi corpus: An annotated corpus with pharmacological substances
  and drug--drug interactions.
\newblock \emph{Journal of biomedical informatics}, 46(5):914--920.

\bibitem[{Kim(2014)}]{DBLP:journals/corr/Kim14f}
Yoon Kim. 2014.
\newblock \href {http://arxiv.org/abs/1408.5882} {Convolutional neural networks
  for sentence classification}.
\newblock \emph{CoRR}, abs/1408.5882.

\bibitem[{Lee et~al.(2020)Lee, Yoon, Kim, Kim, Kim, So, and Kang}]{biobert}
Jinhyuk Lee, Wonjin Yoon, Sungdong Kim, Donghyeon Kim, Sunkyu Kim, Chan~Ho So,
  and Jaewoo Kang. 2020.
\newblock Biobert: a pre-trained biomedical language representation model for
  biomedical text mining.
\newblock \emph{Bioinformatics}, 36(4):1234--1240.

\bibitem[{Li et~al.(2016)Li, Sun, Johnson, Sciaky, Wei, Leaman, Davis,
  Mattingly, Wiegers, and Lu}]{CDR}
Jiao Li, Yueping Sun, Robin~J Johnson, Daniela Sciaky, Chih-Hsuan Wei, Robert
  Leaman, Allan~Peter Davis, Carolyn~J Mattingly, Thomas~C Wiegers, and Zhiyong
  Lu. 2016.
\newblock Biocreative v cdr task corpus: a resource for chemical disease
  relation extraction.
\newblock \emph{Database}, 2016.

\bibitem[{Lin et~al.(2019)Lin, Chen, Chen, and Ren}]{lin2019kagnet}
Bill~Yuchen Lin, Xinyue Chen, Jamin Chen, and Xiang Ren. 2019.
\newblock Kagnet: Knowledge-aware graph networks for commonsense reasoning.
\newblock \emph{arXiv preprint arXiv:1909.02151}.

\bibitem[{Liu et~al.(2019)Liu, Ott, Goyal, Du, Joshi, Chen, Levy, Lewis,
  Zettlemoyer, and Stoyanov}]{liu2019roberta}
Yinhan Liu, Myle Ott, Naman Goyal, Jingfei Du, Mandar Joshi, Danqi Chen, Omer
  Levy, Mike Lewis, Luke Zettlemoyer, and Veselin Stoyanov. 2019.
\newblock Roberta: A robustly optimized bert pretraining approach.
\newblock \emph{arXiv preprint arXiv:1907.11692}.

\bibitem[{Mintz et~al.(2009)Mintz, Bills, Snow, and
  Jurafsky}]{mintz2009distant}
Mike Mintz, Steven Bills, Rion Snow, and Dan Jurafsky. 2009.
\newblock Distant supervision for relation extraction without labeled data.
\newblock In \emph{Proceedings of the Joint Conference of the 47th Annual
  Meeting of the ACL and the 4th International Joint Conference on Natural
  Language Processing of the AFNLP}, pages 1003--1011.

\bibitem[{Soares et~al.(2019)Soares, FitzGerald, Ling, and Kwiatkowski}]{mtb}
Livio~Baldini Soares, Nicholas FitzGerald, Jeffrey Ling, and Tom Kwiatkowski.
  2019.
\newblock Matching the blanks: Distributional similarity for relation learning.
\newblock \emph{arXiv preprint arXiv:1906.03158}.

\bibitem[{Taboureau et~al.(2010)Taboureau, Nielsen, Audouze, Weinhold,
  Edsg{\"a}rd, Roque, Kouskoumvekaki, Bora, Curpan, Jensen et~al.}]{chemprot}
Olivier Taboureau, Sonny~Kim Nielsen, Karine Audouze, Nils Weinhold, Daniel
  Edsg{\"a}rd, Francisco~S Roque, Irene Kouskoumvekaki, Alina Bora, Ramona
  Curpan, Thomas~Sk{\o}t Jensen, et~al. 2010.
\newblock Chemprot: a disease chemical biology database.
\newblock \emph{Nucleic acids research}, 39(suppl\_1):D367--D372.

\bibitem[{Vaswani et~al.(2017)Vaswani, Shazeer, Parmar, Uszkoreit, Jones,
  Gomez, Kaiser, and Polosukhin}]{vaswani2017attention}
Ashish Vaswani, Noam Shazeer, Niki Parmar, Jakob Uszkoreit, Llion Jones,
  Aidan~N Gomez, {\L}ukasz Kaiser, and Illia Polosukhin. 2017.
\newblock Attention is all you need.
\newblock In \emph{Advances in neural information processing systems}, pages
  5998--6008.

\bibitem[{Wu et~al.(2019)Wu, Luo, Leung, Ting, and Lam}]{GDA}
Ye~Wu, Ruibang Luo, Henry~CM Leung, Hing-Fung Ting, and Tak-Wah Lam. 2019.
\newblock Renet: A deep learning approach for extracting gene-disease
  associations from literature.
\newblock In \emph{International Conference on Research in Computational
  Molecular Biology}, pages 272--284. Springer.

\bibitem[{Yao et~al.(2019)Yao, Ye, Li, Han, Lin, Liu, Liu, Huang, Zhou, and
  Sun}]{yao2019docred}
Yuan Yao, Deming Ye, Peng Li, Xu~Han, Yankai Lin, Zhenghao Liu, Zhiyuan Liu,
  Lixin Huang, Jie Zhou, and Maosong Sun. 2019.
\newblock Docred: A large-scale document-level relation extraction dataset.
\newblock \emph{arXiv preprint arXiv:1906.06127}.

\bibitem[{Zelenko et~al.(2003)Zelenko, Aone, and
  Richardella}]{zelenko2003kernel}
Dmitry Zelenko, Chinatsu Aone, and Anthony Richardella. 2003.
\newblock Kernel methods for relation extraction.
\newblock \emph{Journal of machine learning research}, 3(Feb):1083--1106.

\bibitem[{Zhang et~al.(2021)Zhang, Chen, Xie, Deng, Tan, Chen, Huang, Si, and
  Chen}]{DOCUNET}
Ningyu Zhang, Xiang Chen, Xin Xie, Shumin Deng, Chuanqi Tan, Mosha Chen, Fei
  Huang, Luo Si, and Huajun Chen. 2021.
\newblock Document-level relation extraction as semantic segmentation.
\newblock \emph{arXiv preprint arXiv:2106.03618}.

\bibitem[{Zhang et~al.(2017)Zhang, Zhong, Chen, Angeli, and
  Manning}]{zhang2017position}
Yuhao Zhang, Victor Zhong, Danqi Chen, Gabor Angeli, and Christopher~D Manning.
  2017.
\newblock Position-aware attention and supervised data improve slot filling.
\newblock In \emph{Proceedings of the 2017 Conference on Empirical Methods in
  Natural Language Processing}, pages 35--45.

\bibitem[{Zhou and Chen(2021{\natexlab{a}})}]{TEMP}
Wenxuan Zhou and Muhao Chen. 2021{\natexlab{a}}.
\newblock An improved baseline for sentence-level relation extraction.
\newblock \emph{arXiv preprint arXiv:2102.01373}.

\bibitem[{Zhou and Chen(2021{\natexlab{b}})}]{NLLIE}
Wenxuan Zhou and Muhao Chen. 2021{\natexlab{b}}.
\newblock Learning from noisy labels for entity-centric information extraction.
\newblock \emph{arXiv preprint arXiv:2104.08656}.

\bibitem[{Zhou et~al.(2021)Zhou, Huang, Ma, and Huang}]{ATLOP}
Wenxuan Zhou, Kevin Huang, Tengyu Ma, and Jing Huang. 2021.
\newblock Document-level relation extraction with adaptive thresholding and
  localized context pooling.
\newblock In \emph{Proceedings of the AAAI Conference on Artificial
  Intelligence}, volume~35, pages 14612--14620.

\end{thebibliography}
\bibliographystyle{acl_natbib}

\end{document}